\documentclass[10pt,twocolumn,letterpaper]{article}

\usepackage{iccv}
\usepackage{times}
\usepackage{epsfig}
\usepackage{graphicx}
\usepackage{amsmath}
\usepackage{amssymb}
\usepackage{booktabs}
\usepackage{multirow, makecell}
\usepackage{subcaption}
\usepackage{algorithm}
\usepackage{algorithmic}
\usepackage[accsupp]{axessibility}

\newtheorem{assumption}{Assumption}

\DeclareMathOperator*{\argmin}{arg\,min}
\usepackage[dvipsnames]{xcolor}

\usepackage[capitalize]{cleveref}
\crefname{section}{Sec.}{Secs.}
\Crefname{section}{Section}{Sections}
\Crefname{table}{Table}{Tables}
\crefname{table}{Tab.}{Tabs.}
\crefname{algorithm}{Algorithm}{Algorithm}
\crefname{assumption}{Assumption }{Assumption }

\iccvfinalcopy 

\def\iccvPaperID{7352} 

\ificcvfinal\fi

\begin{document}

\title{Efficient Joint Optimization of Layer-Adaptive Weight Pruning \\ in Deep Neural Networks}

\author{Kaixin Xu$^{1,2,\ast}$  \quad Zhe Wang$^{1,2,\ast}$  \quad Xue Geng$^{1}$ \quad Jie Lin$^{1,\dagger}$ \quad Min Wu$^{1}$ \quad Xiaoli Li$^{1,2}$ \quad  Weisi Lin$^{2}$\\
$^{1}$Institute for Infocomm Research (I$^2$R), Agency for Science, Technology and Research (A*STAR),\\1 Fusionopolis Way, 138632, Singapore\\
$^{2}$ Nanyang Technological University, Singapore\\
{\tt\small \{\href{mailto:xuk@i2r.a-star.edu.sg}{xuk},\href{mailto:wangz@i2r.a-star.edu.sg}{wangz},\href{mailto:geng_xue@i2r.a-star.edu.sg}{geng\_xue},\href{mailto:wumin@i2r.a-star.edu.sg}{wumin},\href{mailto:xlli@i2r.a-star.edu.sg}{xlli}\}@i2r.a-star.edu.sg, \href{mailto:jie.dellinger@gmail.com}{jie.dellinger@gmail.com}, \href{mailto:wslin@ntu.edu.sg}{wslin}@ntu.edu.sg}
}

\maketitle
\def\thefootnote{*}\footnotetext{These authors contributed equally to this work}
\def\thefootnote{$\dagger$}\footnotetext{Jie Lin made his contributions when he was with I$^2$R.}

\ificcvfinal\thispagestyle{empty}\fi

\begin{abstract}
In this paper, we propose a novel layer-adaptive weight-pruning approach for Deep Neural Networks (DNNs) that addresses the challenge of optimizing the output distortion minimization while adhering to a target pruning ratio constraint. Our approach takes into account the collective influence of all layers to design a layer-adaptive pruning scheme.
We discover and utilize a very important additivity property of output distortion caused by pruning weights on multiple layers. This property enables us to formulate the pruning as a combinatorial optimization problem and efficiently solve it through dynamic programming. By decomposing the problem into sub-problems, we achieve linear time complexity, making our optimization algorithm fast and feasible to run on CPUs.
Our extensive experiments demonstrate the superiority of our approach over existing methods on the ImageNet and CIFAR-10 datasets. On CIFAR-10, our method achieves remarkable improvements, outperforming others by up to 1.0\% for ResNet-32, 0.5\% for VGG-16, and 0.7\% for DenseNet-121 in terms of top-1 accuracy. On ImageNet, we achieve up to 4.7\% and 4.6\% higher top-1 accuracy compared to other methods for VGG-16 and ResNet-50, respectively. These results highlight the effectiveness and practicality of our approach for enhancing DNN performance through layer-adaptive weight pruning.
Code will be available on \href{https://github.com/Akimoto-Cris/RD\_VIT\_PRUNE}{\url{https://github.com/Akimoto-Cris/RD\_VIT\_PRUNE}}.

\end{abstract}

\section{Introduction}
\label{sec:intro}

\begin{figure*}[t!]
  \centering
  \begin{subfigure}{0.216\linewidth}
    \includegraphics[width=\linewidth]{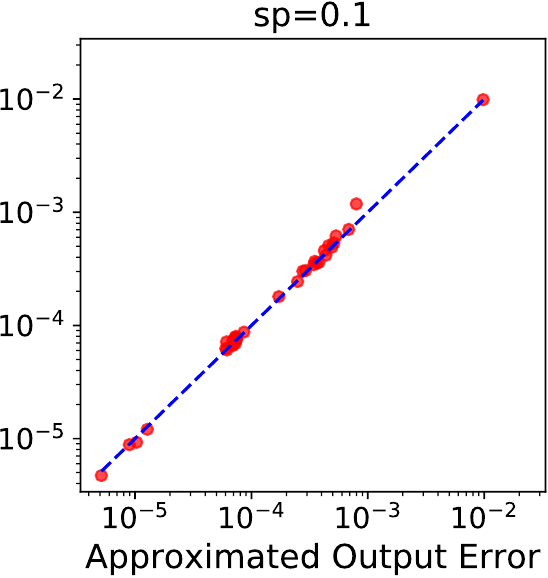}
    \caption{Sparsity=$0.1$.}
    \label{fig:a}
  \end{subfigure}
  \hfill
  \begin{subfigure}{0.24\linewidth}
    \includegraphics[width=\linewidth]{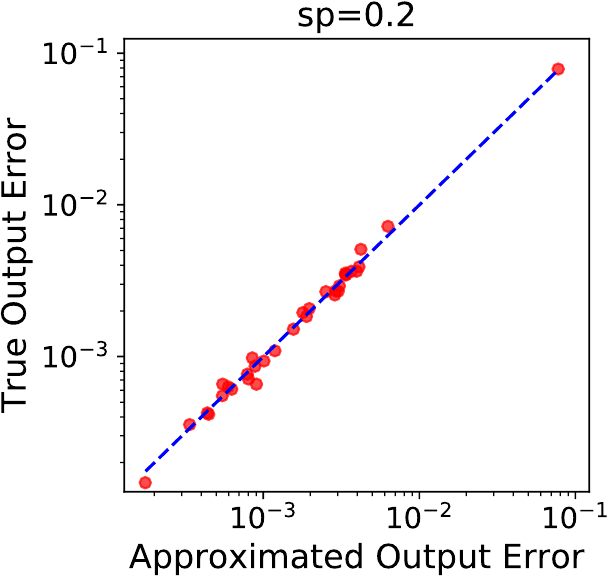}
    \caption{Sparsity=$0.2$.}
    \label{fig:b}
  \end{subfigure}
  \begin{subfigure}{0.234\linewidth}
    \includegraphics[width=\linewidth]{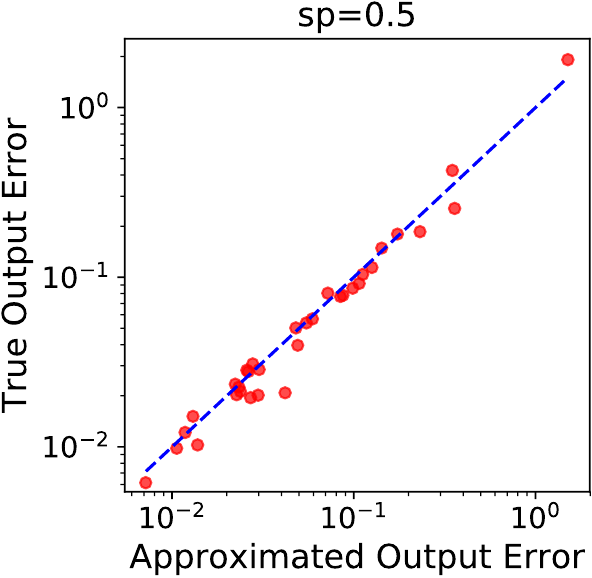}
    \caption{Sparsity=$0.5$.}
    \label{fig:c}
  \end{subfigure}
  \hfill
  \begin{subfigure}{0.244\linewidth}
    \includegraphics[width=\linewidth]{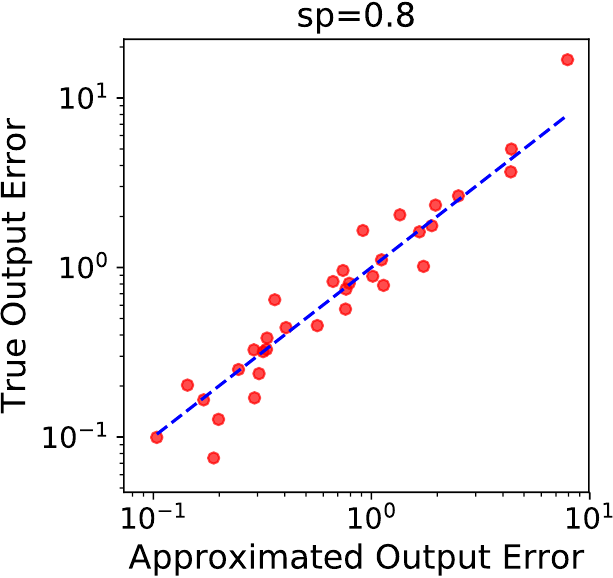}
    \caption{Sparsity=$0.8$.}
    \label{fig:d}
  \end{subfigure}
\caption{An example of additivity property collected on ResNet-32 on CIFAR-10. The vertical axis shows the output distortion when pruning only two consecutive layers. The horizontal axis shows the sum of the output distortion due to the pruning the involved two layers individually. Sub-figures display the situations when all layers in the model are assigned with the corresponding sparsity. }
\label{fig:intro}
\end{figure*}

Deep Neural Networks (DNNs) \cite{Krizhevsky2012nips, vgg, Szegedy2015CVPR, he2016deep, densenet} play a critical role in various computer vision tasks.
However, to achieve high accuracy, DNNs typically require large number of parameters, which makes it very energy-consuming and is difficult to be deployed on resource-limited mobile devices \cite{han2015learning, han16iclr}. 
Pruning is one of the powerful ways to reduce the complexity of DNNs. 
By removing the redundant parameters, the operations can be significantly reduced (e.g., FLOPs), which leads to faster speed and less energy-consuming.
Typically, pruning approaches can be divided into two categories: structured pruning \cite{guo2016dynamic, aghasi2017net, dong2017learning, luo2017thinet, he2017channel, liu2017learning} and weight (unstructured) pruning \cite{li2016pruning, molchanov2016pruning, lin2017runtime, yang2017designing, han2015learning, han16iclr}.
Structured pruning approaches consider a channel or a kernel as a basic pruning unit, while weight pruning approaches consider a weight as a basic pruning unit.
The former is more hardware-friendly and the latter is able to achieve higher pruning ratio.

In this paper, we focus on improving the weight pruning and propose a novel jointly-optimized layer-adaptive approach to achieve state-of-the-art results between FLOPs and accuracy.
Recent discoveries \cite{evci2020rigging, gale2019state, lamp} demonstrate that layer-adaptive sparsity is the superior pruning scheme. 
However, one drawback in prior layer-adaptive approaches is that they only consider the impact of a single layer when  deciding the pruning ratio of that layer. 
The mutual impact between different layers is ignored.
Moreover, another challenge is that the search space of the pruning ratio for each layer increases exponentially as the number of layers.
In a deep neural network, the number of layers can be a hundred or even a thousand, which makes it very difficult to find the solution efficiently.

In our approach, we define a joint learning objective to learn the layer-adaptive pruning scheme. 
We aim to minimize the output distortion of the network when pruning weights on all layers under the constraint of target pruning ratio. 
As the output distortion is highly related to accuracy, our approach is able to maintain accuracy even at high pruning ratios.
We explore an important property of the output distortion and find that the additivity property~\cite{zhe2019optimizing,zhe2021rate,wang2022rdo} holds when we prune weights on multiple layers.
In other words, the output distortion caused by pruning all layers' weights equals to the sum of the output distortion due to the pruning of each individual layer.
We provide a mathematical derivation for the additivity property by using the Taylor series expansion.

Moreover, utilizing the additivity property, we develop a very fast method to solve the optimization via dynamic programming, which has only linear time complexity.
We rewrite the objective function as a combinatorial optimization problem.
By defining the state function and the recursive equation between different states, we can decompose the whole problem into sub-problems and solve it via dynamic programming.
In practice, our approach is able to find the solution in a few minutes on CPUs for deep neural networks.
Note that different from other approximation algorithms, dynamic programming is able to find the global optimal solution, which means that our approach provides optimal pruning scheme with minimal output distortion.
We summarize the main contributions of our paper as follows:

\begin{itemize}
\setlength\itemsep{-0.4em}
\item We propose a novel layer-adaptive pruning scheme that jointly minimizes the output distortion when pruning the weights in all layers. As the output distortion is highly related to the accuracy, our approach maintains high accuracy even when most of the weights are pruned. We also explore an important additivity property for the output distortion based on Taylor series expansion.

\item We develop a fast algorithm to solve the optimization via dynamic programming. The key idea is to rewrite the objective function as a combinatorial optimization problem and then relax the whole problem into tractable sub-problems. Our method can find the solution of a deep neural network in a few minutes.

\item Our approach improves state-of-the-arts on various deep neural networks and datasets.
\end{itemize}

The rest of our paper is organized as follows.
We discuss the related works in section 2. 
In section 3, we develop our approach in detail. We present the objective function, the optimization method, and the time complexity analysis of the algorithm.
In the last section, we provide the comprehensive experimental results.




\section{Related Works}
Our focus of this work generally falls into the magnitude-based pruning (MP) track within model compression of neural networks, with early works such as OBD~\cite{lecun1989optimal}. MP is done by ranking or penalizing weights according to some criterion (\emph{e.g.} magnitude) and removing low-ranked weights. Many efforts have been made ever since under the 
context of~\cite{lecun1989optimal,han2015learning}, which can be roughly divided into the following approaches depending on their timing of pruning embedded in the network training. 

\noindent\textbf{Post-training Pruning.}
Post-training pruning scheme prunes out network parameters after standard network training, \emph{i.e.} prunes from a pretrained converged model. Under this context, parameters can be pruned out at once to achieve target sparsity constraint (ont-shot pruning), or pruned out gradually during the sparse model fine-tuning (iterative pruning). \cite{han2015learning} proposed an iterative pruning scheme that determines layerwise sparsity using layer statistics heuristic. \cite{zhu2018prune,evci2020rigging} adopted a global pruning threshold throughout all layers in the network to meet the model sparsity constraint. 
\cite{carreira2018learning} 
\cite{morcos2019one} pooled all layers together and determined pruning thresholds for different layers in an integrated fashion. 
\cite{frankle2018the} proposed to rewind the weights from previous iterative pruning phase based on the lottery ticket hypothesis.
LAMP\cite{lamp} derived a closed-form layerwise sparsity selection from a relaxed layerwise $l2$ distortion minimization problem that is compatible with various post-training pruning schemes including iterative and one-shot pruning. 
PGMPF~\cite{pgmpf} adopted simple $l2$-based layerwise pruning criterion and improved the weight masking and updating rules during finetuning.
\cite{oto} adopted a one-shot pruning method by leveraging zero-invariant groups.
\cite{lazarevich2021post} proposed to re-calibrate the biases and variances of model weights and activations, similar to the widely adopted bias correction in model quantization~\cite{finkelstein2019fighting,banner2019post}.
\cite{molchanov2016pruning} presented an iterative-pruning method that leverage taylor expansion of model loss and derived a gradient based pruning criteria. Our method leverages taylor expansion on output distortion parametrized by layer weights, which is fundamentally different from ~\cite{molchanov2016pruning}. 
SuRP~\cite{isik2022information} recursively applies triangular inequality and assumes laplacian distribution to approximate output distortion to achieve joint-optimzation similar to us. However, our approximation is more straight-forward and do not need any assumptions on the distribution.

\noindent\textbf{Pruning at Initialization.}
In contrast to the previous scheme, there is an emerging line of work that aims to remove connections or neurons from scratch at the initialization of training, with the merit of avoiding pretraining and complex pruning schedules. 
SNIP~\cite{snip} prunes parameters only once at the initialization phase of training. The normalized magnitudes of the derivatives of parameters are defined as the pruning criterion. 
\cite{jorge2021progressive} presented a modified saliency metric based on SNIP~\cite{snip},  allowing for calculating saliences of partially pruned networks. 
\cite{wang2019picking} engineered the gradient flow when training sparse networks from scratch to achieve better convergence.
Since pruning at initialization out of our research scope, one may refer to related surveys~\cite{wang2021recent} for more comprehensive introduction.

\noindent\textbf{Other Pruning Schemes.}
\cite{bellec2018deep} interleaves the pruning in between normal training course, gradually pruning out more connections and neurons from the networks. This scheme is similar to the previous iterative pruning, however, here the model is trained from scratch. 
ProbMask~\cite{zhou2021effective} similarly leverages projected gradient descent with progressive pruning stretegy to directly train sparse networks. 
\cite{zhang2022optg} integrates supermask training with gradient-drive sparsity for training sparse networks.

Since our main contribution is the improvement of the pruning criteria, we mainly evaluate our method under post-training unstructured pruning paradigms, such as iterative pruning and one-shot pruning. Although our method may have equal potential effectiveness on other sparsity structures and pruning schemes like Pruning at Initialization, we leave such validations for future works. 

\section{Approach}
In this section, we present our approach in detail. 
We first give the formulation of our objective function and then provide the optimization method.
An additivity property is derived based on Taylor series approximation.
The implementation details of the dynamic programming and the analysis of the time complexity are also provided.

\subsection{Objective Function}
Following the notations in~\cite{lamp}, let $f$ denote a neural network, define $W^{(1:l)} = \big(W^{(1)},...,W^{(l)}\big)$ as all the parameters of $f$, where $l$ is the number of layers and $W^{(i)}$ is the weights in layer $i$.
When we prune part of the parameters of $f$, we will receive a modified neural network with the new parameter set $\Tilde{W}^{(1:l)}$.
We view the impact of pruning as the distance between the network outputs $f(x;W^{(1:l)})$ and $f(x;\Tilde{W}^{(1:l)})$.
The learning objective is to minimize the output distortion caused by pruning under the constraint of the pruning ratio,
\begin{eqnarray} \label{eq:obj}
  \begin{split}
 \min \, \|f(x;W^{(1:l)}) - f(x;\Tilde{W}^{(1:l)})\|^{2} \,\,\,\, s.t. \, \frac{\|\Tilde{W}^{(1:l)}\|_0}{\|W^{(1:l)}\|_0} \leq R,
 \end{split}
\end{eqnarray}
where $R$ denotes the pruning ratio for the entire network.


An important property we discover is that the expectation of output distortion, caused by pruning all layers' weights, equals the sum of expectation of output distortion due to the pruning of each individual layer,
\begin{equation}\label{eq:add}
     E\left(\| f(x;W^{(1:l)}) - f(x;\Tilde{W}^{(1:l)}) \|^2\right) = \sum_{i=1}^l E(\delta_{i}),
\end{equation}
where $\delta_{i}$ denotes the output distortion when only pruning the weights in layer $i$.

\subsection{Analysis}
We provide a mathematical derivation for the additivity property. We make the following two assumptions for the proof of additivity property:
\begin{assumption}\label{assum:taylor}
    \textbf{Taylor first order expansion}: The neural network $f$ parametrized by $W^{(1:l)}$ when given a small perturbation $\Delta W^{(1:l)}$ resulting in $\Tilde{W}^{(1:l)}=W^{(1:l)}+\Delta W^{(1:l)}$ can be expanded as the following: 
    \begin{equation} \label{eq:taylor}
    f(x;\Tilde{W}^{(1:l)}) = f(x;W^{(1:l)}) + \sum_{i=1}^l \frac{\partial f}{\partial {W^{(i)}}}  \Delta{W^{(i)}}.
    \end{equation}
\end{assumption}
\begin{assumption}\label{assum:ind-zeromean} 
    \textbf{I.d.d. weight perturbation across layers}~\cite{zhou2018adaptive}: $\forall{0<i\ne j<L}, E(\Delta W^{(i)}) E(\Delta W^{(j)}) = 0$.
\end{assumption}

\noindent According to \cref{eq:taylor}, $\delta = \|f(x;W^{(1:l)}) - f(x;\Tilde{W}^{(1:l)}) \|^2$ can be written as
\begin{align}\label{F2}
\begin{split}
  \delta = \Big(\sum_{i=1}^l {\Delta{W^{(i)}}}^{\top}  {\frac{\partial f}{\partial {W^{(i)}}}}^{\top}  \Big) \Big(\sum_{j=1}^l \frac{\partial f}{\partial {W^{(j)}}} \Delta{W^{(j)}} \Big).
\end{split}
\end{align}
When we take the expectation of \cref{F2} for both sides, the right hand side can be opened up into additive terms (vector transpose is agnostic inside expectation):
    \begin{equation}\label{F3}
    E(\delta) = \sum_{1\le i,j \le l} {E\left(\Delta W^{(i)}\frac{\partial f}{\partial {W^{(i)}}}\right) E\left(\Delta W^{(j)}\frac{\partial f}{\partial W^{(j)}}\right)}.
\end{equation}
Further, since the derivative $\frac{\partial f}{\partial {W^{(i)}}}$ is a constant as we consider trained fixed network weights, we can derive the following from \cref{assum:ind-zeromean}:
\begin{equation}
    E\left(\Delta W^{(i)}\frac{\partial f}{\partial {W^{(i)}}}\right) E\left(\Delta W^{(j)}\frac{\partial f}{\partial {W^{(j)}}}\right) = 0.
\end{equation}
Therfore, the cross terms ($i\ne j$) in \cref{F3} disappear, obtaining:
\begin{align}\label{conc}
\begin{split}
E(\delta) = \sum_{i=1}^l E\left(\|\frac{\partial f}{\partial {W^{(i)}}}  \Delta{W^{(i)}}\|^2\right).
\end{split}
\end{align}

Eq. (\ref{conc}) is the result we want because, again, according to \cref{assum:taylor}, 
\begin{align}
\begin{split}
\frac{\partial f}{\partial {W^{(i)}}}  \Delta{W^{(i)}}  &= f(x;W^{(1:i-1)},\Tilde{W}^{(i)},W^{(i+1,l)}) \\&- f(x;W^{(1;l)}).
\end{split}
\end{align}
Therefore, 
the left hand side of \cref{conc} becomes the real output distortion $\delta$ when all layers are pruned, and the right hand side becomes the sum of the output distortion due to the individual pruning of each single layer's weights, which can be used to approximate the output distortion.

We have done an empirical examination of our theorectically proposed additivity property on real network. As shown in \cref{fig:intro}, when we examine the cases where only pruning two adjacent layers each time in a pretrained model, contributing to the right hand side addable distortion terms while other layers contributing zero to the approximation, we observe that the additivity holds quite well with marginal residuals, where almost all observation points sit close to the identity line.

\begin{algorithm}[t]
\renewcommand{\algorithmicrequire}{\textbf{Input:}}
\renewcommand{\algorithmicensure}{\textbf{Output:}}
\caption{Optimization via dynamic programming. }\label{algo:dp}
\begin{algorithmic}
\REQUIRE Output distortion $\delta_{i}(j)$ when pruning $j$ weights in single layer $i$, for $1 \leq i \leq l$ and $1 \leq j \leq T$. 
\ENSURE The number of weights $p_{i}$ pruned in layer $i$.
\STATE \textbf{Initialize} minimal output distortion $g_{i}^{j} = 0$ when pruning $j$ weights in the first $i$ layers. 
\STATE \textbf{Initialize} state function $s_{i}^{j} = -1$ where $s_{i}^{j}$ denotes the number of weights pruned in layer $i$ when pruning $j$ weights in the first $i$ layers.
\FOR{$i$ from $1$ to $l$}
\FOR{$j$ from $0$ to $T$}
\STATE If $i=1$: $g_{1}^j = \delta_{1}(j)$, $s_{1}^j=j$.
\STATE Else: $g_{i}^{j} = \min \{g_{i-1}^{j-k} + \delta_{i}(k)\}$, $s_{i}^{j} = \argmin_{k} \{g_{i}^{j}\}$.
\ENDFOR
\ENDFOR
\STATE The number of weights pruned in layer $l$ is $p_{l} = s_{l}^{T}$. 
\STATE Update $T=T-s_{l}^{T}$.
\FOR{$i$ from $l-1$ to $1$}
\STATE The number of weights pruned in layer $i$ is $p_{i} = s_{i}^{T}$. 
\STATE Update $T=T-s_{i}^{T}$.
\ENDFOR
\end{algorithmic}
\end{algorithm}

\subsection{Optimization via Dynamic Programming}
By utilizing the additivity property, we can rewrite the objective function as a combinatorial optimization problem and solve it efficiently using dynamic programming.
The objective function is re-written as,
\begin{equation} \label{eq:dp}
 \min \, \delta_{1} + \delta_{2} + ... + \delta_{l} \,\,\,\, s.t. \,\,\,\, t_{1} + t_{2} + ... + t_{l} = T,
\end{equation}
where $T$ denotes the total number of weights to prune and $t_{i}$ denotes the number of weights to prune in layer $i$.
We solve (\ref{eq:dp}) by decomposing the whole problem into sub-problems.
The basic idea is that we define a state function and find the recursive equation between the states.
The problem is solved based on the recursive equation.

Specifically, define $g$ as the state function, in which $g_{i}^{j}$ means the minimal distortion caused when pruning $j$ weights at the first $i$ layers.
Our goal is to calculate $g_{l}^{T}$.
For initialization, we have,
\begin{equation}
g_{1}^{j} = \delta_{1}(j), \,\, for \,\,\, 1 \leq j \leq T,
\end{equation}
where $\delta_{i}(j)$ denotes the distortion caused when pruning $j$ weights at layer $i$.
Then we have the recursive equation between the states $g_{i}$ and $g_{i-1}$, which is,
\begin{equation} \label{eq:transfer}
g_{i}^{j} = \min \{g_{i-1}^{j-k} + \delta_{i}(k)\}, \,\, where \,\,\, 1 \leq k \leq j.
\end{equation}
The state functions are calculated based on equation (\ref{eq:transfer}) in a bottom-up manner from $g_{1}$ to $g_{l}$.
In practice, we need another variable $s$ to store the decision of each state to know the number of weights pruned in each layer. 
$s$ is defined as 
\begin{equation} \label{eq:sij}
s_{i}^{j} = \argmin_{k} \{g_{i}^{j} = g_{i-1}^{j-k} + \delta_{i}(k)\}.
\end{equation}
\cref{algo:dp} shows the pseudo-codes to calculate the state function and find the pruning solution.

\subsection{Time complexity analysis}

The time complexity of the optimization algorithm using dynamic programming is $O(l\times T^{2})$, as we have $l\times T$ different states, and each state needs to enumerate the number of weights pruned in a layer. In practice, this algorithm is very fast which costs just a few seconds on CPUs for deep neural networks. We show the detailed results of the speed in the experimental section.

\begin{table*}[!h]
\centering
\begin{tabular}{c|c|c|cc|cc}
\hline\hline
Dataset                    & Arch                              & Method                & \makecell{Sparsity\\(\%)} & \makecell{Remaining\\FLOPs (\%)} & Top-1 (\%) $\uparrow$ & \makecell{Top-1\\drop (\%)} $\downarrow$ \\ \hline
\multirow{24}{*}{CIFAR-10}  & \multirow{8}{*}{\makecell{ResNet-32\\(Dense: $93.99$)}} & LAMP~\cite{lamp} & $79.03$       & $36.02$    & $92.58\pm0.25$  & $1.41$ \\
                           &                                   & Ours & $58.65$       & $34.5$     & $\mathbf{93.62\pm0.23}$        & $\mathbf{0.37}$ \\ \cline{3-7} 
                           &                                   & LAMP~\cite{lamp} & $89.3$        & $21.66$    & $91.94\pm0.24$        & $2.05$ \\ 
                           &                                   & Ours & $73.76$       & $22.21$    & $\mathbf{93.34\pm0.10}$        & $\mathbf{0.65}$ \\ \cline{3-7} 
                           &                                   & LAMP~\cite{lamp} & $95.5$        & $11$       & $90.04\pm0.13$        & $3.95$ \\ 
                           &                                   & Ours & $86.57$       & $11.25$    & $\mathbf{92.56\pm0.20}$        & $\mathbf{1.43}$ \\ \cline{3-7} 
                           &                                   & LAMP~\cite{lamp} & $98.85$       & $3.29$     & $83.66\pm0.29$        & $10.33$ \\ 
                           &                                   & Ours & $95.5$        & $3.59$     & $\mathbf{90.83\pm0.24}$        & $\mathbf{3.16}$ \\ \cline{2-7} 
                           & \multirow{15}{*}{\makecell{VGG-16\\(Dense: $91.71$)}}    & E-R ker.~\cite{evci2020rigging}  & $95.6$   & /  & $91.99\pm0.14$ & $-0.79$ \\
                           &                                   & DPF~\cite{Lin2020Dynamic} & $95$        & /    & $93.87\pm0.15$        & $-0.13$  \\
                           &                                   & LAMP~\cite{lamp} & $95.6$        & $15.34$    & $92.06\pm0.21$        & $-0.86$  \\
                           &                                   & SuRP~\cite{isik2022information} & $95.6$        & /    & $92.13$        & $-0.93$  \\
                           &                                   & Ours & $95.6$        & $6.83$     & \textcolor{Red}{$\mathbf{92.59\pm0.17}$}        & \textcolor{Red}{$\mathbf{-0.88}$} \\  \cline{3-7}
                           & & Global~\cite{morcos2019one}  & $98.85$ & / & $81.56\pm3.73$ &  $9.64$ \\
                           & & Uniform~\cite{zhu2018prune} & $98.85$ & / & $55.68\pm12.20$ & $35.52$ \\
                           & & Uniform+~\cite{gale2019state} & $98.85$ & / & $87.85\pm0.26$ & $3.35$ \\  
                           & & E-R ker.~\cite{evci2020rigging} & $98.85$       & /          & $90.55\pm0.19$        & $0.65$ \\
                           &                                   & LAMP~\cite{lamp} & $98.85$       & $6.65$     & $91.07\pm0.4$         & $0.13$  \\  
                           &                                   & SuRP~\cite{isik2022information} & $98.84$        & /    & $91.21$        & $-0.01$  \\
                           &                                   & Ours & $98.85$       & $3.43$     & \textcolor{Red}{$\mathbf{92.14\pm0.18}$}        & \textcolor{Red}{$\mathbf{-0.43}$} \\ \cline{3-7} 
                           &                                   & PGMPF~\cite{pgmpf}    & /             & $33$       & $\mathbf{93.6}$                & $0.08$ \\
                           &                                   & LAMP~\cite{lamp} & $86.58$        & $33.53$    & $92.22\pm0.05$        & $-0.51$ \\
                           &                                   & Ours & $67.21$       & $35.49$    & $92.76\pm0.18$        & $\mathbf{-1.05}$ \\ \cline{2-7} 
                           & \multirow{10}{*}{\makecell{DenseNet-121\\(Dense: $91.14$)}}& LAMP~\cite{lamp} & $95.5$        & $6.45$     & $90.11\pm0.13$ & $1.03$ \\ 
                           & & SuRP~\cite{isik2022information} & $95.5$        & /     & $90.75$        & $0.39$ \\ 
                           &                                   & Ours & $95.5$        & $6.72$     & \textcolor{Red}{$\mathbf{91.49\pm0.21}$}        & \textcolor{Red}{$\mathbf{-0.35}$} \\ \cline{3-7} 
                           & & Global~\cite{morcos2019one}  & $98.85$ & / & $45.30\pm27.75$ &  $45.84$ \\
                           & & Uniform~\cite{zhu2018prune} & $98.85$ & / & $66.46\pm18.72$ & $24.68$ \\
                           & & Uniform+~\cite{gale2019state} & $98.85$ & / & $69.25\pm19.28$ & $21.89$ \\ 
                           & & E-R ker.~\cite{evci2020rigging} & $98.85$       & /          & $59.06\pm25.61$        & $32.08$ \\
                           &                                   & LAMP~\cite{lamp} & $98.85$       & $1.71$     & $85.13\pm0.31$        & $6.01$ \\ 
                           &                                   & SuRP~\cite{isik2022information} & $98.56$       & /     & $86.71$        & $4.43$ \\ 
                           &                                   & Ours & $98.85$       & $2.02$     & \textcolor{Red}{$\mathbf{87.7\pm0.24}$}         & \textcolor{Red}{$\mathbf{3.44}$} \\ \hline
\multirow{14}{*}{ImageNet} & \multirow{6}{*}{\makecell{VGG-16-BN\\(Dense: $73.37$)}}    & LAMP~\cite{lamp} & $95.5$        & $37.16$    & $64.63$   & $8.73$  \\ 
                           &                                   & Ours & $95.5$        & $9.12$     & \textcolor{Red}{$\mathbf{66.9}$}                & \textcolor{Red}{$\mathbf{6.47}$} \\ 
                           &                                   & Ours & $73.54$       & $34.95$    & $\mathbf{69.35}$               & $\mathbf{4.02}$ \\ \cline{3-7} 
                           &                                   & LAMP~\cite{lamp} & $98.85$       & $16.73$    & $51.59$               & $21.78$ \\ 
                           &                                   & Ours & $89.3$        & $17.71$    & $\mathbf{68.88}$               & $\mathbf{4.49}$ \\ 
                           &                                   & Ours & $98.85$       & $3.51$     & \textcolor{Red}{$\mathbf{59.41}$}               & \textcolor{Red}{$\mathbf{13.96}$} \\  \cline{2-7}
                           & \multirow{10}{*}{\makecell{ResNet-50\\(Dense: $76.14$)}} & PGMPF~\cite{pgmpf}                & /            & $53.5$     & $75.11$  & $0.52$      \\ 
                           & & Ours & $41$ & $53.5$ & $\mathbf{75.90}$ & $\mathbf{0.24}$ \\\cline{3-7}
                           & & LAMP~\cite{lamp} & $89.3$        & $26.1$     & $72.56$   & $3.58$ \\
                           &                                   & Ours & $67.22$      & $28.52$     & $\mathbf{73.47}$               & $\mathbf{2.67}$ \\ \cline{3-7}
                           &                                   & LAMP~\cite{lamp} & $95.5$        & $15.47$    & $66.04$               & $10.1$ \\ 
                           &                                   & Ours & $95.5$        & $2.85$     & \textcolor{Red}{$\mathbf{66.06}$}               & \textcolor{Red}{$\mathbf{10.08}$} \\
                           &                                   & Ours & $79.01$       & $16.58$    & $\mathbf{72.26}$               & $\mathbf{3.88}$ \\ \cline{3-7}
                           &                                   & LAMP~\cite{lamp} & $98.85$       & $6.15$     & $42.54$               & $33.61$ \\  
                           &                                   & Ours & $91.41$       & $6.07$     & $\mathbf{67.91}$               & $\mathbf{8.23}$ \\\hline\hline
\end{tabular}\vspace{-10pt}
\caption{Iterative Pruning results. \textbf{Bold} denotes the highest Top-1 accuracy or lowest accuracy drop among the results with about the same remaining FLOPs; \textcolor{Red}{\textbf{Red}} denotes the highest Top-1 among that with about the same sparsity. Dense denotes the Top-1 accuracy of the unpruned model.}\vspace{-10pt}
\label{tab:iter_prune}
\end{table*}

\section{Experiment Results}

\begin{figure*}[t!]
  \centering
  \begin{subfigure}{0.49\textwidth}\centering
    \includegraphics[width=\textwidth]{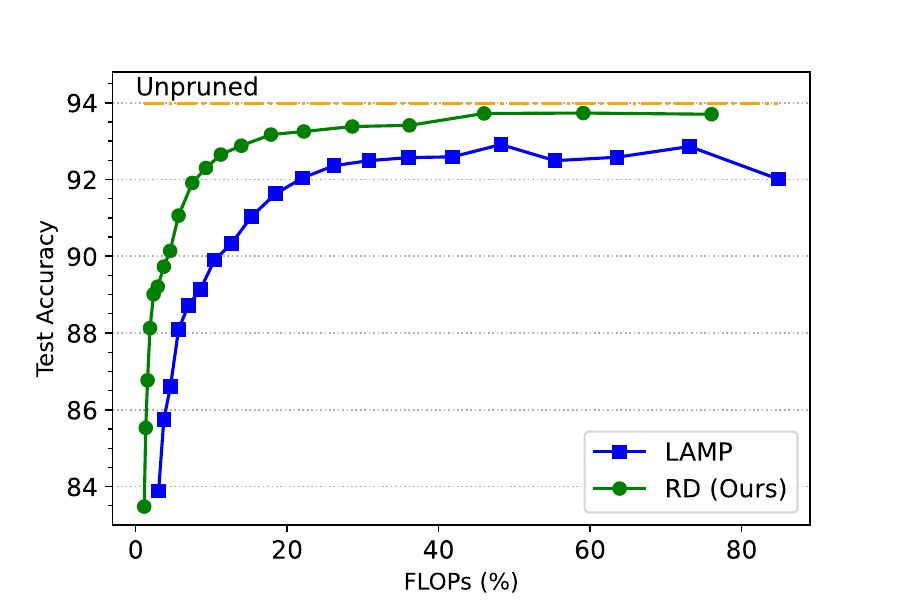}
    \caption{ResNet-32 on CIFAR-10.}
    \label{fig:res32_cif}
  \end{subfigure}
  \begin{subfigure}{0.49\textwidth}\centering
    \includegraphics[width=\textwidth]{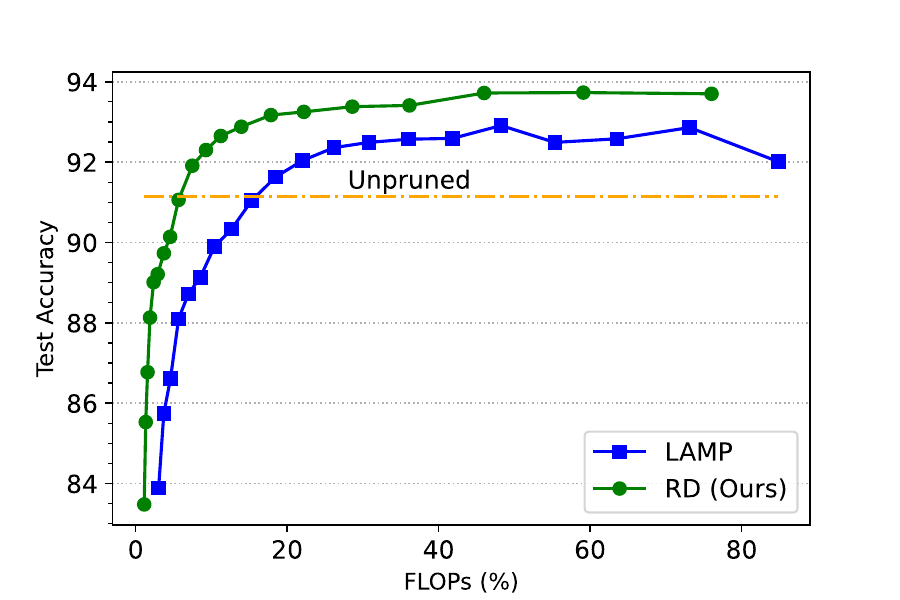}
    \caption{DenseNet-121 on CIFAR-10.}
    \label{fig:dense_cif}
  \end{subfigure}
  \begin{subfigure}{0.49\textwidth}\centering
    \includegraphics[width=\textwidth]{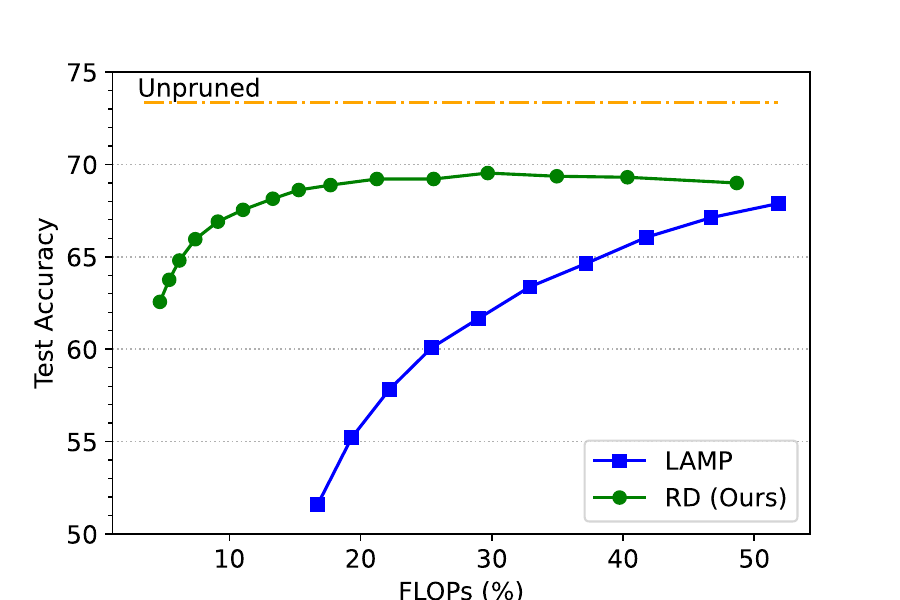}
    \caption{VGG-16 on ImageNet.}
    \label{fig:vgg_img}
  \end{subfigure}
  \begin{subfigure}{0.49\textwidth}\centering
    \includegraphics[width=\textwidth]{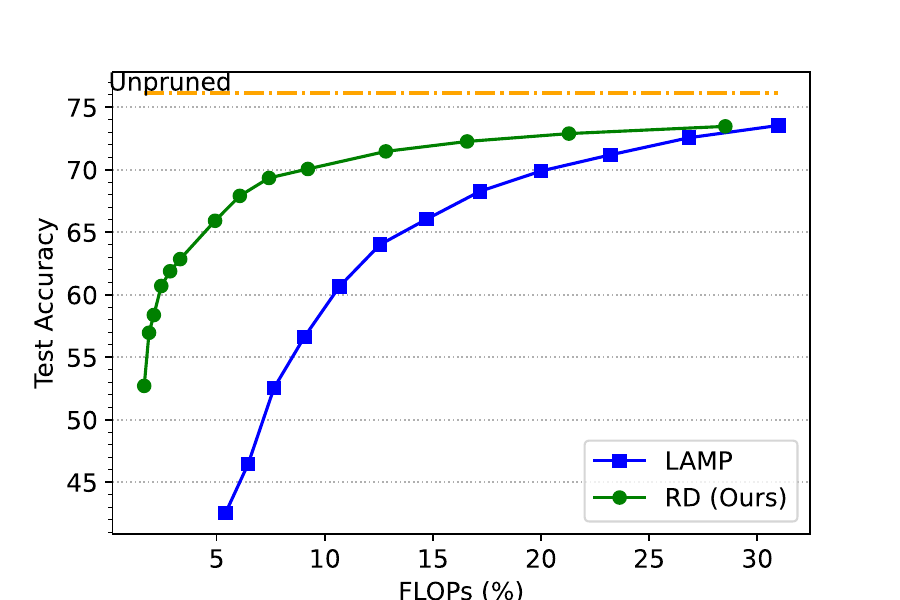}
    \caption{ResNet-50 on ImageNet.}
    \label{fig:res50_img}
  \end{subfigure}\vspace{-10pt}
  \caption{Iterative pruning process of various classification models and datasets.}\vspace{-10pt}
  \label{fig:iter_prune}
\end{figure*}
\noindent\textbf{Implementation Details.}
As our contribution to the existing pruning schemes is on the layer-wise sparsity selection, we evaluate our rate-distortion-based pruning method under different experimental settings, including iterative pruning and one-shot pruning, as well as on multiple network architectures and image classification datasets. We consider 3 models on CIFAR-10 dataset~\cite{cifar}, \emph{i.e.}, VGG-16 following the adapted architectures in~\cite{lamp}, ResNet-32~\cite{he2016deep}, DenseNet-121~\cite{densenet}, while on ImageNet dataset~\cite{imagenet}, we evaluate VGG-16 with BatchNorm~\cite{vgg} and ResNet-50~\cite{he2016deep}. 
On CIFAR-10, following the baseline method~\cite{lamp}, we perform \textit{five independent trials} for each method, and we report the averages and standard deviations among the trials. On the much larger scaled ImageNet, we only perform one trial for each method.
For other implementation details, please refer to the supplementary material.

\noindent\textbf{Details when generating rate-distortion curves.}
In the experimentations, we need to generate rate-distortion curves for every layer to enable sparsity optimization, where points on the curves are a pair of sparsity level and the model output distortion when certain layer is pruned to that sparsity. For non-data-free scheme, the curves are sampled on a randomly selected calibration set from training dataset, while it is also possible to make it data-free by leveraging synthesized data sampled from certain distribution, \emph{e.g.} standard normal distribution. The size of calibration set is set to $1024$ samples for CIFAR-10 and $256$ for ImageNet respectively. However, rate-distortion curves obtained by the above process may be interfered by real-world factors resulting in noisy curves. Therefore we designed various strategies to refine the raw rate-distortion curves and better aid the optimization thereafter. 
Specifically, (1) \textbf{Worst case sampling}: inspired by LAMP~\cite{lamp}, we calculate the distortion as the \emph{maximum} squared norm error among all calibration samples instead of calculating the \emph{MSE} for the whole calibration set; (2) \textbf{Outliers filtering}: treat the local maxima points on the curves that break monotonicity as outlier noises and remove them in order to facilitate \cref{algo:dp}, especially to effectively perform \cref{eq:sij}. We provide ablation studies in the later \cref{sec:abl} to discuss the individual effects of these strategies.

\subsection{Iterative Pruning Results}
In the iterative pruning scheme, one starts with a pretrained full-capacity model. During the finetuning process of the pretrained model, we gradually prune out parameters from the model by a certain amount at each iterative stage. Under different stages of iterative pruning, we get a set of sparse models with gradually increasing sparsity and decreasing computation complexity (FLOPs). 
Following the iterative pruning settings in LAMP~\cite{lamp}, we prune out $20\%$ of the remaining parameters from the model each time after a round of finetuning. The hyper-parameters setup of the finetuning is detailed in the supplementary material.
\cref{tab:iter_prune} compares the results of model accuracies produced during the iterative pruning process by our method and other pruning method counterpart. 

Given non-standardized adoption of CNN models for experimentations in post-training pruning works, we examined as most models as appeared in various literature and add those as baselines in our comparison, including Global~\cite{morcos2019one}, Uniform~\cite{zhu2018prune}, Uniform+~\cite{gale2019state}, LAMP~\cite{lamp}, E-R ker.~\cite{evci2020rigging}, which is an extended Erdős-Rényi method for CNNs pruning, where layer-wise sparsity is selected by a closed-form criterion dependent on merely the layer architecture (e.g., the numbers of input and output channels, the convolutional kernel sizes). 
\cref{fig:iter_prune} further demonstrates the detailed iterative pruning procedures of different methods, where the remaining FLOPs (X-axis) gradually decreases in the course of finetuning.

\noindent\textbf{Results on CIFAR.}
From \cref{tab:iter_prune}, we observe that our method consistently produces pruned models with higher test performance and less test accuracy drop for the same computational complexity (FLOPs) compared to other methods. \cref{fig:iter_prune} further verifies that this observation holds throughout the pruning procedure. 
For example for ResNet-32 on CIFAR-10, our method obtains Top-1 accuracy at $92.56$ on average with $11.25\%$ remaining FLOPs, while baseline result~\cite{lamp} is only $90.04$ at $11\%$ FLOPs; When remaining FLOPs is around $3\%$, we even improve the accuracy by $7.17$, \emph{i.e.} only $3.16$ accuracy drop with only $4.4\%$ survived parameters. 
For VGG-16 on CIFAR-10, we also observe similar results, where our method achieves the least accuracy drop among various counterparts, \emph{e.g.}, when FLOPs are within the range of $33\pm2\%$, without the advance design of soft gradient masking and weight updating strategies adopted in ~\cite{pgmpf}, ours achieves $-1.05\%$ drop of Top-1 at $35.49\%$ FLOPs, which means that the pruned network performs better by $1.05\%$ than the unpruned one. PGMPF~\cite{pgmpf} achieves a higher accuracy score than us on VGG-16 model with $33\%$ remaining FLOPs, which was obtained from a higher performance unpruned model, but still underperforms us regarding the accuracy drop (Top-1 dropped by $0.08\%$).

\noindent\textbf{Results on ImageNet.}
On the larger scale dataset ImageNet, we also observe similar behaviors from our approach. 
For VGG16-BN, we outperform others on both $35\pm2\%$ and $16\pm2$ FLOPs groups. Noticeably, when model sparsity is as high as $98.85\%$, \emph{i.e.} only $1.15\%$ surviving parameters, our method still has $59.41\%$ accuracy, while LAMP already drops to around $52$. 
This is also observed on ResNet-50, where we outperform LAMP by a large margin at $6\%$ FLOPs group. 
From \cref{fig:vgg_img}, there is a minor observation that although consistently higher test accuracy with $<50\%$ FLOPs, VGG-16-BN performs slightly lower within the $30~50\%$ FLOPs range before going up again in the following finetuning iterations. It is speculated that VGG-16-BN is more sensitive to large structural changes for post-train pruning. 

In all, for both datasets, we observe that our method generates higher accuracy sparse models given either the same FLOPs or sparsity constraint.

\subsection{One-shot Pruning Results}
\begin{table}[!h]
    \centering
    \begin{tabular}{c|cc|cc}\hline\hline
        Method & \makecell{Sparsity\\(\%)} & \makecell{Remaining\\FLOPs (\%)} & \makecell{Top-1\\(\%)} & \makecell{Top-1\\drop(\%)} \\ \toprule
        Unpruned & $0$ & $100$ & $76.14$ & - \\\midrule
        LAMP~\cite{lamp}   & $64.5$        & $55$                 & $75.43$    & $0.71$                               \\ 
        OTO~\cite{oto}    & $64.5$        & $34.5$               & $75.1$     & $1.04$                               \\ 
        Ours   & $58$          & $34.5$               & $\mathbf{75.59}$    & $\mathbf{0.55}$                               \\ \hline\hline
    \end{tabular}
  \caption{One-shot pruning results of ResNet-50 on ImageNet.} 
  \label{tab:oneshot}
\end{table}
In one-shot pruning scheme, we directly prune the model to the target computation or parameter constraint, followed by a one-time finetuning. 
\cref{tab:oneshot} summarizes the one-shot pruning results using various unstructured pruning algorithms. We carry out comparison on ResNet-50 on ImageNet. The result verifies that our method still fits in the one-shot pruning scheme, with higher accuracy at $34.5\%$ FLOPs than both baselines~\cite{lamp,oto}.

\subsection{Zero-data Pruning}
\begin{table}[!h]
    \centering
    \begin{tabular}{c|cc|cc}\hline\hline
        Method & \makecell{Sparsity\\(\%)} & \makecell{Remaining\\FLOPs (\%)} & \makecell{Top-1\\(\%)} & \makecell{Top-1\\drop(\%)} \\ \toprule
        Unpruned & $0$ & $100$ & $76.14$ & - \\\midrule
        \cite{lazarevich2021post}    & $50$        & / & $73.89$     & $2.16$            \\ 
        LAMP~\cite{lamp}    & $50$        & $67.05$  &  $74.9$ &  $1.24$                 \\
        Ours*   & $50$        & $42.48$              & $\mathbf{75.13}$    & $\mathbf{1.01}$        \\ \hline\hline
    \end{tabular}
  \caption{Zero-data one-shot pruning results of ResNet-50 on ImageNet dataset. \textbf{Ours*} denotes the zero-data alternative of our method by using white noise data to generate rate-distortion curves. } 
  \label{tab:zerodata}
\end{table}

To evaluate whether our method is compatible with zero-data pruning scheme, which is promising to achieve better generalizability than standard pruning schemes that are usually data dependant, we attempt to adopt our method to zero-data pruning, by replacing the calibration images set that is sampled from real test set with white noise (pixels in each color channel are independently generated by the same distribution required by the classification model, \emph{e.g.}, standard normalized distribution $\mathcal{N}(0, 1)$).

\cref{tab:zerodata} summarizes the results of zero-data variant of our approach compared to the baseline~\cite{lazarevich2021post} result with the same data synthesize strategy (white-noise). We also include LAMP~\cite{lamp} in the comparison since it happens to require no calibration set to obtain layerwise pruning thresholds for good. Our approach still achieves superior results under zero-data scenario, with only $1.01\%$ performance drop. This is within our expectation since our rate-distortion theory-based algorithm does not depend on any specific input data distribution.

\begin{figure*}[!h]
  \centering
  \begin{subfigure}{0.45\linewidth}\centering
    \includegraphics[width=.9\linewidth]{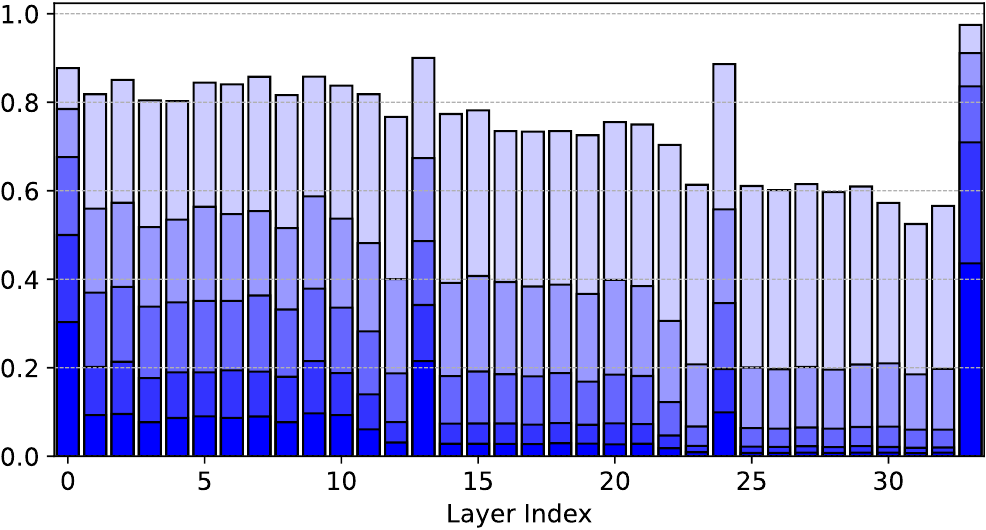}
    \caption{LAMP.}
    \label{fig:lamp_sp}
  \end{subfigure}
  \hfill
  \begin{subfigure}{0.45\linewidth}\centering
    \includegraphics[width=.9\linewidth]{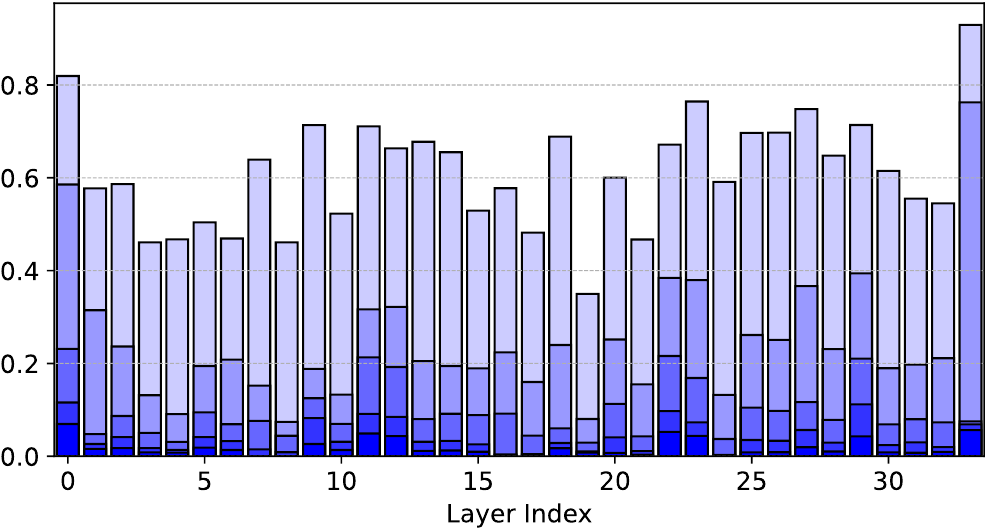}
    \caption{Ours.}
    \label{fig:rd_sp}
  \end{subfigure}
  \caption{Layer-wise sparsity statistics of ResNet-32 on CIFAR-10 of different methods during iterative pruning. Height of bars denotes the pruning rate with $\{0.36, 0.74, 0.89, 0.96, 0.98\}$ model sparsities.}
  \label{fig:sp_distribution}
\end{figure*}

\subsection{Ablation Studies}
\label{sec:abl}

\begin{table}[!h]
\centering
\begin{tabular}{c|c|c|c|c}\hline\hline
Arch                       & Method             & \makecell{Sparsity\\(\%)} & \makecell{Top-1\\(\%)} & \makecell{Top-1\\drop(\%)} \\ \hline
\multirow{2}{*}{ResNet-50} & Ours       & $58$         & $\mathbf{75.59}$    & $\mathbf{0.55}$                               \\ \cline{2-5} 
                           & Ours* & $60$         & $74.89$    & $1.45$                               \\ \hline
\multirow{2}{*}{VGG-16-BN} & Ours       & $60$         & $\mathbf{69.01}$    & $\mathbf{4.36}$                               \\ \cline{2-5} 
                           & Ours* & $59$          & $62.50$    & $10.87$                              \\ \hline\hline
\end{tabular}
\caption{Comparison of joint-optimization objective and vanilla (single-layer) optimization (denoted by Ours*).}
\label{tab:abl_joint}
\end{table}
\begin{table}[th]
    \centering
    \begin{tabular}{cc|c|cc}\hline\hline
        WCS & OF & Sparsity (\%) & \makecell{Top-1\\(\%)} & \makecell{Top-1\\drop(\%)} \\\hline
        \multicolumn{2}{c|}{Unpruned} & $0$ & $93.99$ & - \\\midrule
         & & $89.3$ & $91.15$ & $2.84$ \\
        $\sqrt{}$ & & $89.2$ & $91.31$ & $2.68$ \\
        & $\sqrt{}$ & $89$ & $91.51$ & $2.58$ \\
        $\sqrt{}$ & $\sqrt{}$ & $89.3$ & $\mathbf{92.3}$ & $\mathbf{1.69}$ \\\hline\hline
    \end{tabular}
    \caption{Different post-processing strategies of RD curves on ResNet-32 on CIFAR-10 with iterative pruning scheme. \textbf{WCS}: Worst case sampling, \textbf{OF}: Outlier filtering.}
    \label{tab:abl_cifar}
\end{table}

Since our major contribution is the joint-optimization strategy, we first conducted a comparison with the case not using joint-optimization where we directly solve layer-wise sparsity on the output features of each layer, resulting in the performance shown in \cref{tab:abl_joint}. As indicated in the table, we observe deteriorated performances for such single-layer optimization on both tested models, showing that our joint-optimization strategy is optimal.

\begin{table}[!h]
    \centering
    \begin{tabular}{cc|c|cc}\hline\hline
        WCS & OF & Sparsity (\%) & \makecell{Top-1\\(\%)} & \makecell{Top-1\\drop(\%)} \\\hline
        \multicolumn{2}{c|}{Unpruned} & $0$ & $76.14$ & - \\\midrule
         & & $60$ & $31.22$ & $44.92$ \\
        $\sqrt{}$ & & $60$ & $31.22$ & $44.92$ \\
        & $\sqrt{}$ & $60$ & $38.12$ & $38.02$ \\
        $\sqrt{}$ & $\sqrt{}$ & $60$ & $38.12$ & $38.02$ \\\hline\hline
    \end{tabular}
    \caption{Different post-processing strategies of RD curves on ResNet-50 on ImageNet with one-shot pruning scheme. Test accuracy of the model \textbf{before} finetuning is reported.}
    \label{tab:abl_img}
\end{table}
We also evaluate the individual effectiveness of the aforementioned rate-distortion curves refining strategies. We first perform ablation on CIFAR-10 dataset. From \cref{tab:abl_cifar}, we observe that at the same model sparsity $89\%$, which is relatively high for one-shot pruning scheme, both strategies are shown to work positively for our approach. Therefore, we included both strategies to conduct experiments of main results. 
We also observe the same on ImageNet dataset, as shown in \cref{tab:abl_img}. 
Particularly, Outlier filtering strategy brings slightly more improvement on both CIFAR-10 and ImageNet, where Worst case sampling makes no difference at this particular sparsity target.

\subsection{Other discussions}

There is also an interesting observation from \cref{tab:iter_prune} that with the same model sparsity, our method constantly reduces more FLOPs from the model. To better analyze this phenomenon, we take a closer look into the layerwise sparsity solution given by different approaches. As shown in \cref{fig:sp_distribution}, our method prunes out more parameters from deeper layers than LAMP~\cite{lamp}. Since activations in deeper layers in CNNs usually have more channels and features than shallow layers, pruning out more parameters from deep layers will reduce more operations, resulting in less remaining FLOPs. 
From \cref{fig:sp_distribution}, another observation is that both methods prune out more parameters from the last layer of ResNet-32 which is the fully-connected layer, implying that parameters of the last layer contain large redundancy. 
Meanwhile, we observe that DenseNet-121 on CIFAR-10 does not display the above phenomenon, where our method reduces the same level of FLOPs compared with LAMP under the same sparsity. We elaborate this in the supplementary material.

\subsection{Time Complexity}

\begin{table}[!h]
    \centering
    \begin{tabular}{c|c|c|c}\hline\hline
        Configuration &\makecell{No.\\layers} & \makecell{Sparsity\\(\%)} & Time (s) \\\hline
        ResNet-32@CIFAR-10 & $35$ & $20$ & $0.46\pm 0.09$ \\\hline
        ResNet-50@ImageNet & $54$ & $50$ & $2.08\pm 0.21$ \\\hline\hline
    \end{tabular}
    \caption{Time spent on layerwise sparsity optimization of our method.}
    \label{tab:time}
\end{table}

We provide the empirical optimization time complexity analysis in \cref{tab:time}. In practice, we use ternary search algorithm to search the solution of $s_i^j$ in \cref{eq:sij}, which has logarithmic time complexity given the search range. On small datasets like CIFAR-10, with $35$ layers, our method takes less than a second to calculate the layerwise sparsity, while on larger ImageNet, our method still only takes a few seconds. 

\begin{table}[!h]
    \centering
    \begin{tabular}{c|c|c}\hline\hline
        Configuration & \makecell{Curve\\Generation (s)} & Optimize (s) \\\hline
        ResNet-18@CIFAR-10 & 1052.64 & 0.84 \\\hline
        VGG-16@CIFAR-10 & 664.19 & 2.20 \\\hline\hline
    \end{tabular}
    \caption{Comparison of time cost of RD curve generations and optimization.}\label{tab:curvetime}
\end{table}

We also  analyze the curve generation costs.
For each layer, we traverse all calibration set samples to calculate output distortion at all sparsity levels to generate a rate-distoriton curve. Therefore, the cost of generating rate-distortion curves becomes $O(lSN)$, where $l$ is the number of layers, $S$ is the number of sparsity levels (we set $S=100$ in practice), and $N$ is the size of calibration set. We provide the actual time costs for two CIFAR-10 models in Tab.~\ref{tab:curvetime}. In practice, we used optimized dataloader and parallelized curve generation of different layers to cut down the inference time per sample.

\subsection{Analysis of Approximation Error}
\begin{figure}[!h]
  \centering
  \includegraphics[width=0.8\linewidth]{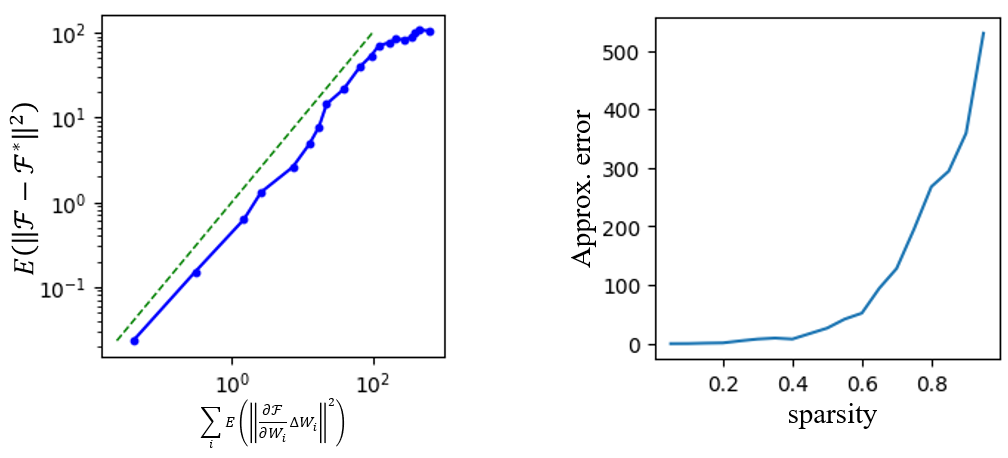}
  \vspace{-10pt}
  \caption{Empirical Approximation Error Analysis.}
  \vspace{-10pt}
  \label{fig:approx_error}
\end{figure}

Given the locality nature of taylor expansion, we expect an increasing discrepancy of the taylor approximation under large distortion. 
We analyze the empirical approximation error in Fig.~\ref{fig:approx_error}. The left figure visualizes the relations between the taylor-based approximated output distortion (X-axis) and the real output distortion (Y-axis), we notice that the data points in the figure are very close to the diagonal. The right figure plots the approximation error at different sparsity levels. The approximation error inflates at large sparsities, e.g. $>50\%$. 

\section{Conclusions}

We have presented a new rate-distortion based unstructured pruning criterion. We revealed the output distortion additivity of CNN models unstructured pruning, supported by theory and experiments. We exploited this property to simplify the NP-hard layerwise sparsity optimization problem into a fast pruning criterion with only $O(l\times T^2)$ complexity. Benefiting from the direct optimization on the output distortion, our proposed criterion shows superiority over existing methods in various post-training pruning schemes. Our criterion prefer to prune deep and large layers, leading to significant model size and FLOPs reductions.

\section*{Acknowledgement}
This research is supported by the Agency for Science, Technology and Research (A*STAR) under its Funds (Project Number A1892b0026 and C211118009 and MTC Programmatic Funds (Grant No. M23L7b0021)). Any opinions, findings and conclusions or recommendations ex- pressed in this material are those of the author(s) and do not reflect the views of the A*STAR.

{\small
\bibliographystyle{ieee_fullname}
\bibliography{egbib}
}

\end{document}